\PassOptionsToPackage{unicode}{hyperref}
\PassOptionsToPackage{hyphens}{url}
\PassOptionsToPackage{dvipsnames,svgnames,x11names}{xcolor}
\documentclass[
]{article}

\usepackage{amsmath,amssymb}
\usepackage{iftex}
\ifPDFTeX
  \usepackage[T1]{fontenc}
  \usepackage[utf8]{inputenc}
  \usepackage{textcomp} 
\else 
  \usepackage{unicode-math}
  \defaultfontfeatures{Scale=MatchLowercase}
  \defaultfontfeatures[\rmfamily]{Ligatures=TeX,Scale=1}
\fi
\usepackage{lmodern}
\ifPDFTeX\else  
\fi
\IfFileExists{upquote.sty}{\usepackage{upquote}}{}
\IfFileExists{microtype.sty}{
  \usepackage[]{microtype}
  \UseMicrotypeSet[protrusion]{basicmath} 
}{}
\makeatletter
\@ifundefined{KOMAClassName}{
  \IfFileExists{parskip.sty}{%
    \usepackage{parskip}
  }{
    \setlength{\parindent}{0pt}
    \setlength{\parskip}{6pt plus 2pt minus 1pt}}
}{
  \KOMAoptions{parskip=half}}
\makeatother
\usepackage{xcolor}
\ifx\paragraph\undefined\else
  \let\oldparagraph\paragraph
  \renewcommand{\paragraph}[1]{\oldparagraph{#1}\mbox{}}
\fi
\ifx\subparagraph\undefined\else
  \let\oldsubparagraph\subparagraph
  \renewcommand{\subparagraph}[1]{\oldsubparagraph{#1}\mbox{}}
\fi

\providecommand{\tightlist}{%
  \setlength{\itemsep}{0pt}\setlength{\parskip}{0pt}}\usepackage{longtable,booktabs,array}
\usepackage{calc} 
\usepackage{etoolbox}
\makeatletter
\patchcmd\longtable{\par}{\if@noskipsec\mbox{}\fi\par}{}{}
\makeatother
\IfFileExists{footnotehyper.sty}{\usepackage{footnotehyper}}{\usepackage{footnote}}
\makesavenoteenv{longtable}
\usepackage{graphicx}
\makeatletter
\def\maxwidth{\ifdim\Gin@nat@width>\linewidth\linewidth\else\Gin@nat@width\fi}
\def\maxheight{\ifdim\Gin@nat@height>\textheight\textheight\else\Gin@nat@height\fi}
\makeatother
\setkeys{Gin}{width=\maxwidth,height=\maxheight,keepaspectratio}
\makeatletter
\def\fps@figure{htbp}
\makeatother
\newlength{\cslhangindent}
\newlength{\csllabelwidth}
\newlength{\cslentryspacingunit} 
\newenvironment{CSLReferences}[2] 
 {
  \setlength{\parindent}{0pt}
  \ifodd #1
  \let\oldpar\par
  \def\par{\hangindent=\cslhangindent\oldpar}
  \fi
  \setlength{\parskip}{#2\cslentryspacingunit}
 }%
 {}
\usepackage{calc}

\usepackage{arxiv}
\usepackage{orcidlink}
\usepackage{amsmath}
\usepackage[T1]{fontenc}
\makeatletter
\@ifpackageloaded{caption}{}{\usepackage{caption}}
\AtBeginDocument{%
\ifdefined\contentsname
  \renewcommand*\contentsname{Table of contents}
\else
  \newcommand\contentsname{Table of contents}
\fi
\ifdefined\listfigurename
  \renewcommand*\listfigurename{List of Figures}
\else
  \newcommand\listfigurename{List of Figures}
\fi
\ifdefined\listtablename
  \renewcommand*\listtablename{List of Tables}
\else
  \newcommand\listtablename{List of Tables}
\fi
\ifdefined\figurename
  \renewcommand*\figurename{Figure}
\else
  \newcommand\figurename{Figure}
\fi
\ifdefined\tablename
  \renewcommand*\tablename{Table}
\else
  \newcommand\tablename{Table}
\fi
}
\@ifpackageloaded{float}{}{\usepackage{float}}
\floatstyle{ruled}
\@ifundefined{c@chapter}{\newfloat{codelisting}{h}{lop}}{\newfloat{codelisting}{h}{lop}[chapter]}
\floatname{codelisting}{Listing}

\makeatother
\makeatletter
\@ifpackageloaded{caption}{}{\usepackage{caption}}
\@ifpackageloaded{subcaption}{}{\usepackage{subcaption}}
\makeatother
\makeatletter
\@ifpackageloaded{tcolorbox}{}{\usepackage[skins,breakable]{tcolorbox}}
\makeatother
\makeatletter
\@ifundefined{shadecolor}{\definecolor{shadecolor}{rgb}{.97, .97, .97}}
\makeatother
\ifLuaTeX
  \usepackage{selnolig}  
\fi
\IfFileExists{bookmark.sty}{\usepackage{bookmark}}{\usepackage{hyperref}}
\IfFileExists{xurl.sty}{\usepackage{xurl}}{} 
\hypersetup{
  pdftitle={Large language models for aspect-based sentiment analysis},
  pdfauthor={Paul F. Simmering; Paavo Huoviala},
  pdfkeywords={aspect-based sentiment analysis (ABSA), large language
models (LLMs), GPT, few-shot learning, prompt engineering},
  colorlinks=true,
  linkcolor={blue},
  filecolor={Maroon},
  citecolor={Blue},
  urlcolor={Blue},
  pdfcreator={LaTeX via pandoc}}

\newcommand{\runninghead}{A Preprint }
\renewcommand{\runninghead}{A Preprint }
\title{Large language models for aspect-based sentiment analysis}
\author{
\textbf{Paul F. Simmering}\\\\Q Agentur für Forschung
GmbH\\\\\href{mailto:paul.simmering@teamq.de}{paul.simmering@teamq.de}\\\\\\
\textbf{Paavo Huoviala}~\orcidlink{0000-0003-2402-304X}\\\\Q Agentur für
Forschung
GmbH\\\\\href{mailto:paavo.huoviala@teamq.de}{paavo.huoviala@teamq.de}}
\date{}
\begin{document}
\maketitle
\begin{abstract}
Large language models (LLMs) offer unprecedented text completion
capabilities. As general models, they can fulfill a wide range of roles,
including those of more specialized models. We assess the performance of
GPT-4 and GPT-3.5 in zero shot, few shot and fine-tuned settings on the
aspect-based sentiment analysis (ABSA) task. Fine-tuned GPT-3.5 achieves
a state-of-the-art F1 score of 83.8 on the joint aspect term extraction
and polarity classification task of the SemEval-2014 Task 4, improving
upon InstructABSA (Scaria et al. 2023) by 5.7\%. However, this comes at
the price of 1000 times more model parameters and thus increased
inference cost. We discuss the the cost-performance trade-offs of
different models, and analyze the typical errors that they make. Our
results also indicate that detailed prompts improve performance in
zero-shot and few-shot settings but are not necessary for fine-tuned
models. This evidence is relevant for practioners that are faced with
the choice of prompt engineering versus fine-tuning when using LLMs for
ABSA.
\end{abstract}
{\bfseries \emph Keywords}
\def\sep{\textbullet\ }
aspect-based sentiment analysis (ABSA) \sep large language models
(LLMs) \sep GPT \sep few-shot learning \sep 
prompt engineering

\ifdefined\Shaded\renewenvironment{Shaded}{\begin{tcolorbox}[boxrule=0pt, interior hidden, breakable, borderline west={3pt}{0pt}{shadecolor}, enhanced, frame hidden, sharp corners]}{\end{tcolorbox}}\fi

\hypertarget{introduction}{%
\section{Introduction}\label{introduction}}

Aspect-based sentiment analysis (ABSA) is used for providing insights
into digitized texts, such as product reviews or forum discussions, and
is therefore a key capability for fields such as digital social
sciences, humanities, and market research. It offers a more detailed
view of reviews compared to conventional sentiment analysis, which
typically categorizes the overall sentiment of a whole text as positive,
negative, or neutral (Turney 2002; Pang, Lee, and Vaithyanathan 2002).
The SemEval-2014 challenge (Pontiki et al. 2014) proposed an aspect term
extraction (ATE) and an aspect term sentiment classification (ATSC)
tasks. These can be merged into a joint task, where the goal is to
simultaneously extract aspect terms and classify their polarity. We
focus our efforts on the joint task in the present study \footnote{Code
  available at \url{https://github.com/qagentur/absa_llm}}.

A wide array of specialized models have been developed for ABSA. With
the emergence of pre-trained language models like BERT (Devlin et al.
2019), the field has witnessed significant advancements in accuracy,
particularly in transformer-based models that incorporate task-specific
output layers. Zhang et al. (2022) provide a comprehensive survey of
similar models in the domain. PyABSA (Yang and Li 2023) is notable for
its collection of ABSA datasets and models for various ABSA tasks and
across multiple languages.

OpenAI has reshaped the field of Natural Language Processing (NLP) in
the recent years with their family of Generative Pre-trained Transformer
models, GPTs (Brown et al. 2020; Ouyang et al. 2022; OpenAI 2023). GPTs
are generalist large language models (LLMs) with a very wide range of
demonstrated capabilities ranging from classic NLP tasks to causal
reasoning, and even functional knowledge in specialized domains like
medicine or law (Mao et al. 2023). Zhang et al. (2023) investigated the
accuracy of LLMs for sentiment analysis and found non-finetuned LLMs
generally capable, but not on par with specialized models, on complex
tasks such as ABSA. They also emphasize that the choice of prompt is
critical for performance.

InstructABSA is the current state of the art for ABSA. It utilizes a a
fine-tuned T5 model (Wang et al. 2022) along with prompt instructions
for improved performance.

To investigate how well do the current generation of LLMs are able to
perform ABSA, we test the performance of GPT-3.5 and GPT-4 (pinned model
gpt-3.5-turbo-0613 and gpt-4-0613 available through OpenAI API) using a
range of conditions, and compare their performance to that of
InstructABSA, the state-of-the-art model for the SemEval-2014 Task 4
Joint Task. \textbf{Table 1} contains an overview of the tested models.
The selection leaves out some notable high performance LLMs, such as
Meta's Llama2 (Touvron et al. 2023a), Anthropic's Claude2, and Google's
Bard.

\begin{longtable}[]{@{}
  >{\raggedright\arraybackslash}p{(\columnwidth - 8\tabcolsep) * \real{0.2000}}
  >{\raggedright\arraybackslash}p{(\columnwidth - 8\tabcolsep) * \real{0.2000}}
  >{\raggedright\arraybackslash}p{(\columnwidth - 8\tabcolsep) * \real{0.2000}}
  >{\raggedright\arraybackslash}p{(\columnwidth - 8\tabcolsep) * \real{0.2000}}
  >{\raggedright\arraybackslash}p{(\columnwidth - 8\tabcolsep) * \real{0.2000}}@{}}
\caption{Model comparison}\tabularnewline
\toprule\noalign{}
\begin{minipage}[b]{\linewidth}\raggedright
\end{minipage} & \begin{minipage}[b]{\linewidth}\raggedright
GPT-4
\end{minipage} & \begin{minipage}[b]{\linewidth}\raggedright
GPT-3.5
\end{minipage} & \begin{minipage}[b]{\linewidth}\raggedright
GPT-3.5 fine tuned
\end{minipage} & \begin{minipage}[b]{\linewidth}\raggedright
InstructABSA
\end{minipage} \\
\midrule\noalign{}
\endfirsthead
\toprule\noalign{}
\begin{minipage}[b]{\linewidth}\raggedright
\end{minipage} & \begin{minipage}[b]{\linewidth}\raggedright
GPT-4
\end{minipage} & \begin{minipage}[b]{\linewidth}\raggedright
GPT-3.5
\end{minipage} & \begin{minipage}[b]{\linewidth}\raggedright
GPT-3.5 fine tuned
\end{minipage} & \begin{minipage}[b]{\linewidth}\raggedright
InstructABSA
\end{minipage} \\
\midrule\noalign{}
\endhead
\bottomrule\noalign{}
\endlastfoot
Base model & GPT-4 & GPT-3.5 & GPT-3.5 & T5 \\
Parameters & \textasciitilde1.7T\footnote{The number of parameters of
  GPT-4 is currently not disclosed by OpenAI. Internet rumors put it at
  1.7T. We use this number for comparison purposes.} & 175B & 175B &
200M \\
Fine-tuning & Not available & No & SemEval-2014 & SemEval2014 \\
Language & Multilingual & Multilingual & English & English \\
Availability & Commercial & Commercial & Commercial, custom & Open
source \\
\end{longtable}

\hypertarget{methods-data}{%
\section{Methods \& Data}\label{methods-data}}

We evaluate performance on the gold standard benchmark dataset
SemEval2014 (Pontiki et al. 2014), consisting of human annotated laptop
and restaurant reviews. Model performance is measured on a joint aspect
term extraction and polarity classification task with using F1 score as
the main performance metric. We test multiple prompt variations, the
number of provided in-context examples, and fine-tune the models via the
OpenAI's API. \textbf{Figure 1} shows the overview of the experimental
set-up. We also break down the types of False positive errors that the
models make in order to get a better understanding on their strengths
and weaknesses, and compare costs and cost-efficiency of the different
options.

\begin{figure}

{\centering \includegraphics{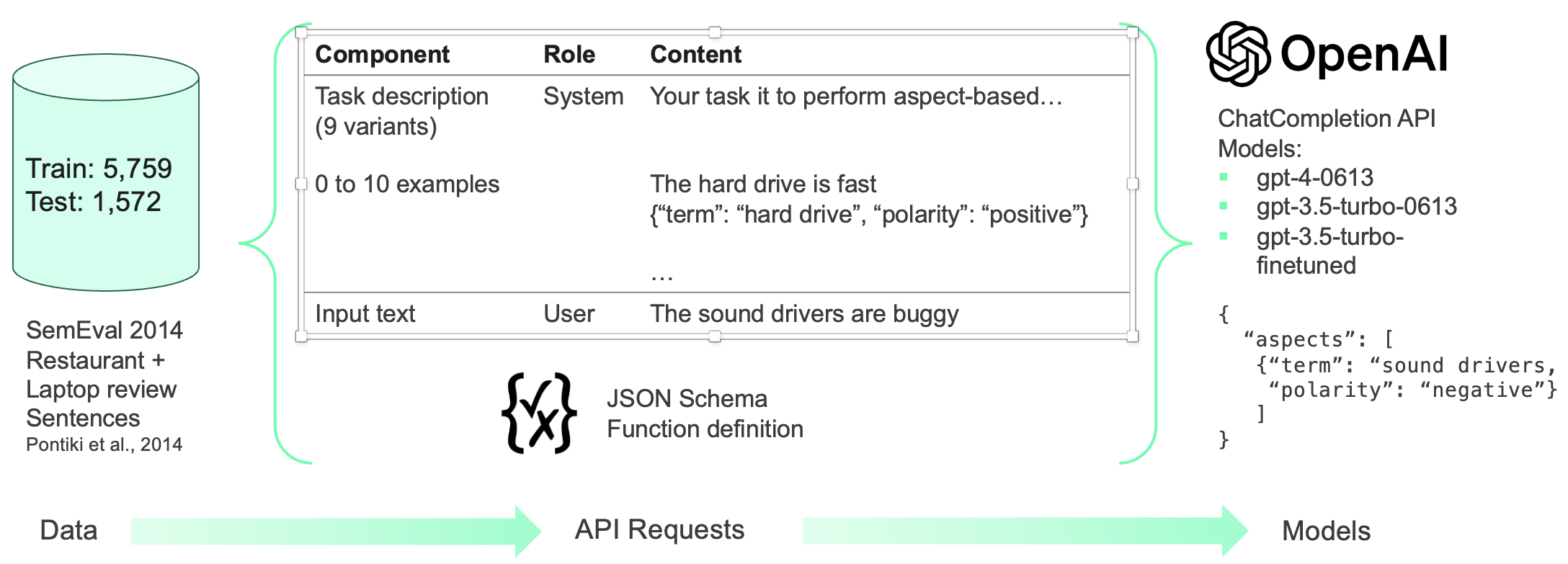}

}

\caption{Overview of experimental setup}

\end{figure}

\hypertarget{data}{%
\subsection{Data}\label{data}}

The SemEval-2014 dataset contains human-annotated sentences from laptop
and restaurant reviews. The original 2014 dataset has 4 polarity labels:
positive, negative, neutral and conflict. To maintain consistency with
InstructABSA and previous models, we also exclude examples with the
``conflict'' label. The examples that include them were removed
completely, rather than just removing the ``conflict'' labeled aspects,
to avoid the case where the text contains aspects that the model is not
supposed to label. Altogether, 126 examples were removed from the
training set and 28 from the test set, leaving 5759 in training and 1572
in test.

We used up to 10 examples from the training set for in-context
instructions and the whole training set for fine-tuning. During
fine-tuning, we reserved 20\% of the training set for validation. We
used the test set to evaluate the final model performance.

\hypertarget{prompts}{%
\subsection{Prompts}\label{prompts}}

The prompts for this study are structured in 3 parts: 1) a task
description in the system message 2) and optional a set of in-context
examples also included in the system message 3) a JSON schema defining
the expected output format.

\hypertarget{task-description}{%
\subsubsection{Task description}\label{task-description}}

The task description is a short text that instructs the model.
Altogether, we used 9 different task descriptions (see \textbf{Table 2};
see Appendix for full prompt texts). They primarily consisted of
variations of the original annotation guidelines for the SemEval2014
task and the prompt used in the state-of-the-art model, InstructABSA. In
addition, we also added a task description to attempt to coax the
generalist GPT model to behave as a specialist model (\emph{Roleplay}),
and to perform the task in a step-wise manner (\emph{Separate tasks}).
Finally, we added two controls: a minimal one sentence summary of the
task (\emph{Minimal}), and a \emph{No prompt} control that was only
tested on the fine-tuned models.

The \emph{Reference} task description explicitly taps into GPT-4's
pre-existing knowledge of SemEval-2014 by referencing the task by name.

The \emph{Guidelines summary} was created by GPT-4 itself. We pasted the
original annotation guidelines into the OpenAI API playground and asked
the model to summarize them. The resulting summary was then used as the
task description.

\hypertarget{in-context-examples}{%
\subsubsection{In-context examples}\label{in-context-examples}}

The examples provided as part of the completion request enable
in-context learning without changing the model's parameters (Liu et al.
2021). Further, they introduce the output format to the model. Providing
examples can have a large positive impact on model performance of
text-to-text models in classification tasks and Min et al. (2022) show
that the main benefit stems from establishing the answer format and
providing information on the distribution of tokens.

We tested a range of in-context example conditions. The examples were
manually picked from the training set based on the curriculum learning
(Bengio et al. 2009) concept, meaning that a series of examples starts
with the simplest case and builds to complex edge cases. We tested the
following conditions, each building on the previous:

\begin{itemize}
\tightlist
\item
  0 shot
\item
  2 shot: one basic example per review domain (restaurants and laptops)
  showing typical case
\item
  4 shot: two basic examples per domain, show variety of outputs
\item
  6 shot: three examples per domain
\item
  10 shot: same as above, plus hard examples for each domain to teach
  edge cases
\end{itemize}

Full texts are provided in the appendix.

\hypertarget{chat-interface}{%
\subsubsection{Chat interface}\label{chat-interface}}

In contrast to prior models, GPT-3.5 and GPT-4 use a chat interface
rather than a pure text-to-text interface. The chat starts with a system
message, which can be used for providing the task description. The model
then predicts an answer and the resulting format is a dialogue between
user messages and assistant responses.

This presents two options for including in-context examples for the
task. Either they are included in the system message, or they are
presented as pairs of user and assistant messages already in the
dialogue. We tested both options in a preliminary experiment and found
that GPT-3.5's performance benefits only from the examples in the system
message. For example, for the \emph{Guidelines summary} prompt
increasing the number of in-context examples from 0 to 6 increased the
F1 score from 64.3 to 65.7 when the examples were provided within the
system message, while same examples actually decreased F1 score from
64.3 to 60.0 when the examples were provided outside of the system
message. GPT-4's performance appears to benefit from both options, but
we chose to include the examples in the system message for both models.

\hypertarget{function-schema}{%
\subsection{Function schema}\label{function-schema}}

As the GPT models are text-to-text models, having a standardized output
format is crucial for its usefulness in a structured task like ABSA.
Prior studies by Zhang et al. (2023) and Scaria et al. (2023) have
devised custom formats and instructed the model to use them in the
prompt. We opted for JSON as a standard and use OpenAI's \emph{function
calling} feature to enforce the format: As part of the inference
command, the model is instructed to call a function with the generated
text as input. This function describes the expected output format using
a JSON schema, standardizing the model outputs. The JSON schema used has
description fields and informative field names, which act as further
in-context instructions.

Given an input text ``The fajitas are tasty'' the schema would then
become:

\begin{verbatim}
"input": "The fajitas are tasty"
"output": {"aspects": [{"aspect": "fajitas", "polarity": "positive"}]}
\end{verbatim}

\hypertarget{fine-tuning-gpt-3.5}{%
\subsection{Fine tuning GPT-3.5}\label{fine-tuning-gpt-3.5}}

We fine-tuned GPT-3.5 on the examples from the training set for 3 epochs
using an 80/20 train/validation split. A system message prompt can be
provided for fine-tuning and for optimal performance. The same system
message should be included at inference time. To test the influence of
the system prompt, we fine-tuned the model with three different prompt:
\emph{Guidelines summary}, \emph{Minimal} system message, and without a
system message (\emph{No prompt}).

\begin{figure}

{\centering \includegraphics{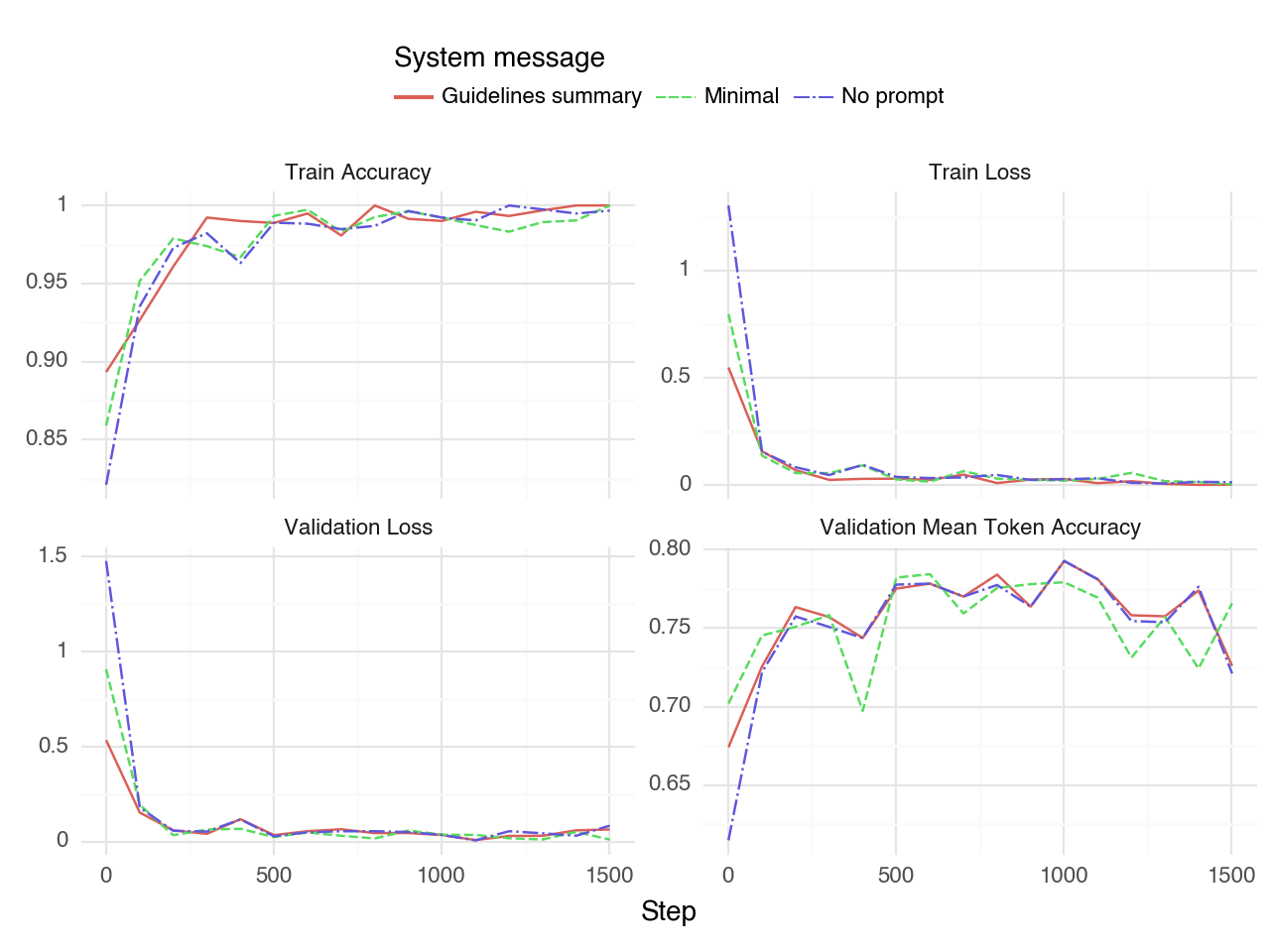}

}

\caption{Fine-tuning}

\end{figure}

\begin{verbatim}
\end{verbatim}

Training converged quickly in all cases (\textbf{Figure 2}). Further
epochs are unlikely to improve performance because a training accuracy
of 100\% was reached. Validation accuracy fluctuates around 76\%.

The resulting fine-tuned models do not need a JSON schema to produce
structured output, as they have learned the expected JSON format.
Parsing the returned strings as JSON did not produce any errors. This
reduces the number of input tokens required for inference.

\hypertarget{parameters}{%
\subsection{Parameters}\label{parameters}}

We iterated on the number of in-context examples and prompt variations.
Temperature was set to 0 to get near-deterministic results. Other
parameters were left at their default values.

\hypertarget{results}{%
\section{Results}\label{results}}

\hypertarget{prompts-1}{%
\subsection{Prompts}\label{prompts-1}}

As prompt selection and tuning is a key step for successful use of a
task-agnostic text-to-text model like GPT, we first wanted to find out
which prompt variants gave the best results in the task. We tested the
prompt variations using GPT-3.5. Results are summarised in \textbf{Table
2}. Overall, the prompts gave fairly similar results and adding
in-context examples did not improve performance significantly, with the
exceptions of the \emph{InstructABSA} and \emph{Minimal} prompts. Using
6 in-context examples was better than 10 for almost all tested prompts.
Examples 7-10 were purposefully selected to represent difficult cases
and apparently failed to be useful for the task. The \emph{Guidelines
summary} provided the best average performance (F1: 64.99) across the
tested in-context examples conditions, by a narrow margin. Overall,
GPT-3.5 (without fine-tuning) did not attain the performance levels of
modern specialized models for ABSA.

\begin{longtable}[]{@{}
  >{\raggedright\arraybackslash}p{(\columnwidth - 10\tabcolsep) * \real{0.1842}}
  >{\raggedleft\arraybackslash}p{(\columnwidth - 10\tabcolsep) * \real{0.1118}}
  >{\raggedright\arraybackslash}p{(\columnwidth - 10\tabcolsep) * \real{0.3421}}
  >{\raggedleft\arraybackslash}p{(\columnwidth - 10\tabcolsep) * \real{0.1184}}
  >{\raggedleft\arraybackslash}p{(\columnwidth - 10\tabcolsep) * \real{0.1184}}
  >{\raggedleft\arraybackslash}p{(\columnwidth - 10\tabcolsep) * \real{0.1250}}@{}}
\caption{Task description variants (GPT-3.5)}\tabularnewline
\toprule\noalign{}
\begin{minipage}[b]{\linewidth}\raggedright
Prompt
\end{minipage} & \begin{minipage}[b]{\linewidth}\raggedleft
Prompt tokens
\end{minipage} & \begin{minipage}[b]{\linewidth}\raggedright
Description
\end{minipage} & \begin{minipage}[b]{\linewidth}\raggedleft
F1, 0 examples
\end{minipage} & \begin{minipage}[b]{\linewidth}\raggedleft
F1, 6 examples
\end{minipage} & \begin{minipage}[b]{\linewidth}\raggedleft
F1, 10 examples
\end{minipage} \\
\midrule\noalign{}
\endfirsthead
\toprule\noalign{}
\begin{minipage}[b]{\linewidth}\raggedright
Prompt
\end{minipage} & \begin{minipage}[b]{\linewidth}\raggedleft
Prompt tokens
\end{minipage} & \begin{minipage}[b]{\linewidth}\raggedright
Description
\end{minipage} & \begin{minipage}[b]{\linewidth}\raggedleft
F1, 0 examples
\end{minipage} & \begin{minipage}[b]{\linewidth}\raggedleft
F1, 6 examples
\end{minipage} & \begin{minipage}[b]{\linewidth}\raggedleft
F1, 10 examples
\end{minipage} \\
\midrule\noalign{}
\endhead
\bottomrule\noalign{}
\endlastfoot
Guidelines summary & 178 & Summary of the guidelines created by GPT-4 &
64.39 & 65.65 & 64.92 \\
Annotation guidelines & 2021 & Official guide for SemEval-2014 Task 4 &
63.77 & 66.03 & 64.88 \\
Roleplay & 84 & Pretend to be a specialized machine learning model &
62.09 & 64.29 & 64.64 \\
Reference & 39 & Name-drop SemEval-2014 Task 4 & 62.36 & 64.2 & 63.23 \\
InstructABSA & 18 & InstructABSA prompt & 52.94 & 61.39 & 61.32 \\
Separate tasks & 150 & 2 steps: Term extraction, polarity classification
& 61.68 & 62.74 & 61.27 \\
InstructABSA with examples & 249 & InstructABSA prompt + 6 examples from
paper & 61.54 & 62.49 & 60.18 \\
Minimal & 20 & One sentence summary of task. & 46.62 & 60.29 & 59.71 \\
\end{longtable}

\hypertarget{gpt-4-and-fine-tuned-gpt-3.5-model-performance}{%
\subsection{GPT-4 and fine-tuned GPT 3.5 model
performance}\label{gpt-4-and-fine-tuned-gpt-3.5-model-performance}}

Next, we conducted experiments using an updated model, GPT-4. We chose
the \emph{Guidelines summary} prompt for this, as it was the best
performing prompt on GPT-3.5. Surprisingly, in the zero-shot condition,
GPT-4 exhibited slightly lower performance compared to GPT-3.5,
achieving an F1 score of 57.6. However, its performance notably improved
with the inclusion of in-context examples, surpassing GPT-3.5 in the 6
and 10 example conditions, reaching a peak F1 score of 71.3.

In contrast, the three fine-tuned GPT-3.5 models consistently
outperformed the previous state-of-the-art model, InstructABSA (Scaria
et al. 2023), even with no in-context examples provided (\textbf{Figure
3}). All three tested models demonstrated remarkably similar
performance. Notably, the \emph{Minimal prompt} achieved the highest F1
score at 83.8, and even the \emph{No prompt} model outperformed the
previous state-of-the-art. The addition of extra in-context examples did
not yield significant performance gains, but rather, even resulted in a
mild performance decrease (tested only on the \emph{Minimal prompt}).
This possibly reflects the mismatch between the system message seen in
fine-tuning and the system message with in-context examples at inference
time. Collectively, these findings suggest that for a relatively
straightforward task like ABSA, optimal performance is achieved by
fine-tuning the model and allowing it to learn the task directly from
the training data.

\begin{figure}

{\centering \includegraphics{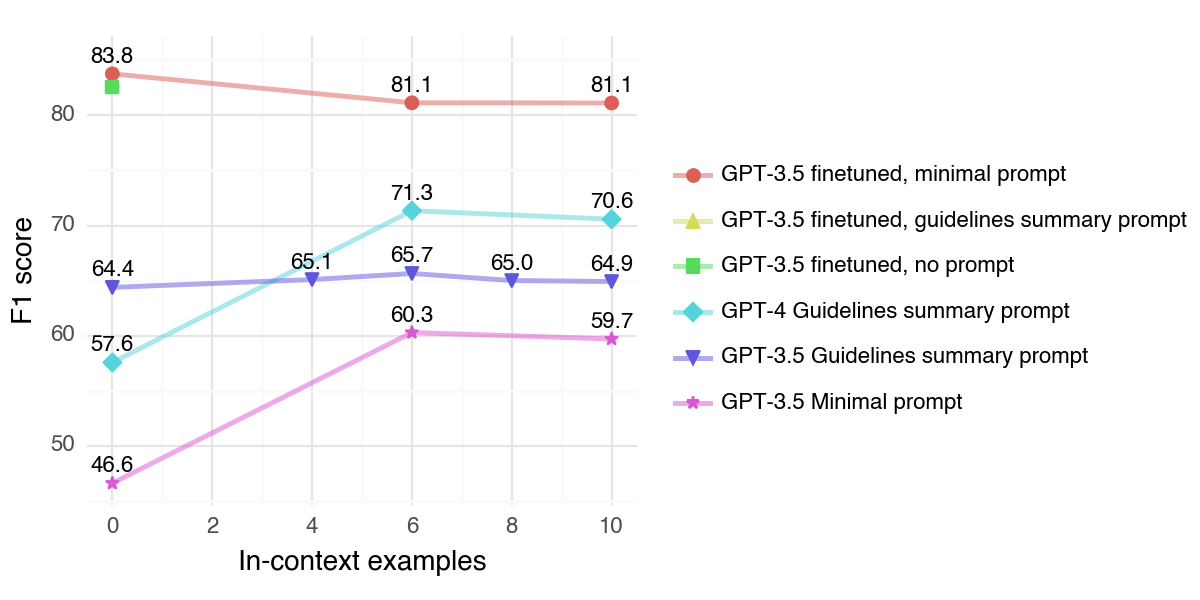}

}

\caption{Model comparison by number of in-context examples}

\end{figure}

\begin{verbatim}
\end{verbatim}

\hypertarget{error-analysis}{%
\subsection{Error Analysis}\label{error-analysis}}

Overall, the fine-tuned GPT-3.5 models are both more sensitive and more
specific than either GPT-3.5 or GPT-4. The tine-tuned GPT-3.5 makes 50\%
less errors than the best GPT-3.5 model, and 42\% less than GPT-4, while
correctly detecting and classifying 34\% (vs.~GPT-3.5) or 21\%
(vs.~GPT-4) more aspect term-polarity pairs than the other tested
models. \textbf{Table 3} shows error metrics for each model.

\begin{longtable}[]{@{}
  >{\raggedright\arraybackslash}p{(\columnwidth - 14\tabcolsep) * \real{0.2599}}
  >{\raggedleft\arraybackslash}p{(\columnwidth - 14\tabcolsep) * \real{0.0339}}
  >{\raggedleft\arraybackslash}p{(\columnwidth - 14\tabcolsep) * \real{0.0339}}
  >{\raggedleft\arraybackslash}p{(\columnwidth - 14\tabcolsep) * \real{0.0339}}
  >{\raggedleft\arraybackslash}p{(\columnwidth - 14\tabcolsep) * \real{0.2203}}
  >{\raggedleft\arraybackslash}p{(\columnwidth - 14\tabcolsep) * \real{0.1695}}
  >{\raggedleft\arraybackslash}p{(\columnwidth - 14\tabcolsep) * \real{0.1356}}
  >{\raggedleft\arraybackslash}p{(\columnwidth - 14\tabcolsep) * \real{0.1130}}@{}}
\caption{Error analysis by model}\tabularnewline
\toprule\noalign{}
\begin{minipage}[b]{\linewidth}\raggedright
Model
\end{minipage} & \begin{minipage}[b]{\linewidth}\raggedleft
TP
\end{minipage} & \begin{minipage}[b]{\linewidth}\raggedleft
FN
\end{minipage} & \begin{minipage}[b]{\linewidth}\raggedleft
FP
\end{minipage} & \begin{minipage}[b]{\linewidth}\raggedleft
FP:Predicted aspect not in gold set
\end{minipage} & \begin{minipage}[b]{\linewidth}\raggedleft
FP:Polarity classification
\end{minipage} & \begin{minipage}[b]{\linewidth}\raggedleft
FP:Aspect boundaries
\end{minipage} & \begin{minipage}[b]{\linewidth}\raggedleft
FP:Made up terms
\end{minipage} \\
\midrule\noalign{}
\endfirsthead
\toprule\noalign{}
\begin{minipage}[b]{\linewidth}\raggedright
Model
\end{minipage} & \begin{minipage}[b]{\linewidth}\raggedleft
TP
\end{minipage} & \begin{minipage}[b]{\linewidth}\raggedleft
FN
\end{minipage} & \begin{minipage}[b]{\linewidth}\raggedleft
FP
\end{minipage} & \begin{minipage}[b]{\linewidth}\raggedleft
FP:Predicted aspect not in gold set
\end{minipage} & \begin{minipage}[b]{\linewidth}\raggedleft
FP:Polarity classification
\end{minipage} & \begin{minipage}[b]{\linewidth}\raggedleft
FP:Aspect boundaries
\end{minipage} & \begin{minipage}[b]{\linewidth}\raggedleft
FP:Made up terms
\end{minipage} \\
\midrule\noalign{}
\endhead
\bottomrule\noalign{}
\endlastfoot
GPT-3.5 & 1426 & 791 & 676 & 346 & 142 & 154 & 34 \\
GPT-3.5 finetuned, guidelines summary prompt & 1907 & 377 & 397 & 81 &
189 & 125 & 2 \\
GPT-3.5 finetuned, minimal prompt & 1909 & 373 & 367 & 68 & 166 & 131 &
2 \\
GPT-3.5 finetuned, no prompt & 1876 & 404 & 387 & 61 & 192 & 132 & 2 \\
GPT-4 & 1576 & 514 & 752 & 415 & 174 & 154 & 9 \\
\end{longtable}

In order to find out what kinds of errors the models make, we also broke
down the false positives further into the following error sub-types:

\begin{itemize}
\tightlist
\item
  \emph{Predicted aspect not in gold set}: The model extracts a term
  from the text that is not found in the gold aspect set.
\item
  \emph{Polarity classification}: The model correctly extracts an aspect
  term but misclassifies its polarity.
\item
  \emph{Aspect boundaries}: The model partially extracts an aspect term
  that has more or fewer tokens than the gold aspect term.
\item
  \emph{Made up terms}: The model predicts an aspect term that is not
  found in the text.
\end{itemize}

In aggregate, the most common error sub-type is \emph{Predicted aspect
not in gold set}. These were often terms that could be of interest in a
real-world use-case, but broke some annotation rules of the benchmark
dataset. For example, labeling non-noun terms in a sentence like:
\emph{``It's fast {[}positive{]}, quiet {[}positive{]}, incredibly small
{[}positive{]} and affordable {[}positive{]}.''} Fine-tuning the model
had also the biggest effect here, leading to up to \textasciitilde88\%
reduction of this type of FPs.

The second most common error sub-type is the \emph{Polarity
classification} error. For example, in the sentence \emph{``The MAC Mini
{[}positive{]}, wireless keyboard / mouse {[}positive{]} and a HDMI
cable {[}positive{]} is all I need to get some real work done.''}
GPT-3.5 labeled multiple of the listed features of the computer as
positive. However, these were annotated as neutral aspects in the gold
set, as no specific positive qualities about the features were
explicitly mentioned in the review. Overall, we observed that all of the
models over-predicted the number of positive and negative labels and
under-predicted the number of neutral labels. It is also worth noting
that the \emph{polarity classification} errors increase with model
performance, as a prerequisite for polarity classification is that the
aspect term is extracted correctly.

The third most common are \emph{Aspect boundary} issues. For example, in
the sentence \emph{``The Mini's body hasn't changed since late 2010- and
for a good reason.''} GPT-4 extracted the term ``Mini's body'' whereas
only the term ``body'' is labelled in the gold set.

Finally, we also occasionally saw the models identifying terms that were
not present in the text. These tended to be abstractions of the sentence
content, such labeling ``speed'' and ``performance'' in the sentence
\emph{``It is extremely fast and never lags''}. GPT-3.5 has a notably
high number of made up terms, seemingly failing to learn the instruction
that the aspect terms must be extracted verbatim. However, altogether
this error sub-types were not that common and finetuning almost
completely eliminated the issue.

On the whole, the errors are typically related to the idiosyncrasies of
the benchmark dataset. In a few-shot setting, the LLM struggles to pick
up on the nuances of labeling rules, instead delivering more commonsense
labels. This ties into remarks by Zhang et al. (2023) who found that
LLMs are capable of ABSA, but not with the precise format required by
the benchmark dataset. While such errors hamper benchmark performance,
they should not necessarily discourage from using LLMs in real-world
applications of ABSA or similar tasks. In domains like market research
it may be preferable to for example also extract non-noun and abstracted
terms,

\hypertarget{economic-analysis}{%
\subsection{Economic analysis}\label{economic-analysis}}

LLMs are computationally expensive to train and use. With OpenAI users
pay based on both the number of input and output tokens\footnote{https://openai.com/pricing}.
OpenAI charges eight times more for input and output tokens for
fine-tuned GPT-3.5 models than for the default model. However,
fine-tuned models do not require the use of a JSON schema, reducing the
number of input tokens required. As demonstrated by our results,
fine-tuned models are also not reliant on relatively long prompts or
presence of in-context examples. Thus, they can lead to cost savings and
providing more accurate results with a lower overall cost. See
\textbf{Table 4} for the overall cost summary of the tested model
versions and conditions.

\begin{longtable}[]{@{}
  >{\raggedright\arraybackslash}p{(\columnwidth - 12\tabcolsep) * \real{0.1638}}
  >{\raggedright\arraybackslash}p{(\columnwidth - 12\tabcolsep) * \real{0.1810}}
  >{\raggedright\arraybackslash}p{(\columnwidth - 12\tabcolsep) * \real{0.1293}}
  >{\raggedleft\arraybackslash}p{(\columnwidth - 12\tabcolsep) * \real{0.1983}}
  >{\raggedleft\arraybackslash}p{(\columnwidth - 12\tabcolsep) * \real{0.1034}}
  >{\raggedleft\arraybackslash}p{(\columnwidth - 12\tabcolsep) * \real{0.1207}}
  >{\raggedleft\arraybackslash}p{(\columnwidth - 12\tabcolsep) * \real{0.1034}}@{}}
\caption{Cost comparison}\tabularnewline
\toprule\noalign{}
\begin{minipage}[b]{\linewidth}\raggedright
Model
\end{minipage} & \begin{minipage}[b]{\linewidth}\raggedright
Prompt
\end{minipage} & \begin{minipage}[b]{\linewidth}\raggedright
JSON schema
\end{minipage} & \begin{minipage}[b]{\linewidth}\raggedleft
In-context examples
\end{minipage} & \begin{minipage}[b]{\linewidth}\raggedleft
F1 score
\end{minipage} & \begin{minipage}[b]{\linewidth}\raggedleft
Cost (USD)
\end{minipage} & \begin{minipage}[b]{\linewidth}\raggedleft
F1 / USD
\end{minipage} \\
\midrule\noalign{}
\endfirsthead
\toprule\noalign{}
\begin{minipage}[b]{\linewidth}\raggedright
Model
\end{minipage} & \begin{minipage}[b]{\linewidth}\raggedright
Prompt
\end{minipage} & \begin{minipage}[b]{\linewidth}\raggedright
JSON schema
\end{minipage} & \begin{minipage}[b]{\linewidth}\raggedleft
In-context examples
\end{minipage} & \begin{minipage}[b]{\linewidth}\raggedleft
F1 score
\end{minipage} & \begin{minipage}[b]{\linewidth}\raggedleft
Cost (USD)
\end{minipage} & \begin{minipage}[b]{\linewidth}\raggedleft
F1 / USD
\end{minipage} \\
\midrule\noalign{}
\endhead
\bottomrule\noalign{}
\endlastfoot
GPT-3.5 finetuned & Minimal & False & 0 & 83.76 & 0.59 & 141.97 \\
GPT-3.5 finetuned & Guidelines summary & False & 0 & 83.13 & 2.08 &
39.97 \\
GPT-3.5 finetuned & No prompt & False & 0 & 82.59 & 0.36 & 229.42 \\
GPT-3.5 finetuned & Minimal & False & 6 & 81.14 & 3.17 & 25.6 \\
InstructABSA & InstructABSA prompt & False & 6 & 79.3 & 0.05 & 1586 \\
GPT-4 & Guidelines summary & True & 6 & 71.34 & 15.02 & 4.75 \\
GPT-3.5 & Guidelines summary & True & 6 & 65.65 & 0.71 & 92.46 \\
GPT-3.5 & Guidelines summary & True & 0 & 64.39 & 0.39 & 165.1 \\
GPT-3.5 & Minimal & True & 6 & 60.29 & 0.54 & 111.65 \\
GPT-4 & Guidelines summary & True & 0 & 57.58 & 8.91 & 6.46 \\
GPT-3.5 & Minimal & True & 0 & 46.62 & 0.24 & 194.25 \\
\end{longtable}

In real-world applications with significantly larger datasets than the
benchmark set used here, it is worth considering that InstructABSA is
still significantly cheaper to operate, while providing near
state-of-the-art results, with a run amounting to less than \$0.05 when
executed on a single vCPU on AWS or a similar cloud provider. GPT-4 on
the other hand, is the strongest model available in a low computational
resource setting and no access to training data for fine-tuning, but
also by far the most expensive model to operate, reflecting its high
parameter count. Notably, when measuring cost-efficiency as a ratio of
obtained F1 score divided by the run cost, InstructABSA is more than
300-fold better than the best performing GPT-4 model, but only
\textasciitilde7-fold better than the most efficient fine-tuned GPT-3.5
model.

\hypertarget{discussion}{%
\section{Discussion}\label{discussion}}

We explore the application of OpenAI's LLMs for the classic NLP task of
aspect based sentiment analysis (ABSA). We focus on the joint aspect
term extraction and polarity classification task on the widely used
SemEval-2014 benchmark dataset. A fine-tuned GPT-3.5 model achieved
state-of-the-art performance on the ABSA joint task (83.8\% F1 score) on
the benchmark task. Fine-tuning the model also emerged as the most
efficient option for the task, offering superior performance without the
need for extensive prompt engineering. This not only saves time but also
reduces token consumption, both valuable aspects in real-world
applications.

Our analysis revealed that the errors made by the not-fine tuned models
were often related to discrepancies between the model's predictions and
the idiosyncrasies of the benchmark dataset's annotation rules. GPT-3.5
and 4 offered often sensible term-polarity predictions that just failed
to take the intricacies of all the annotation rules into account,
whereas fine-tuning seemed to align the models better with this specific
formulation of the ABSA task. This is supported by how the main
performance increase between the basic and fine-tuned models seemed to
result both from decreased number of False negatives as well as near
90\% reduction in False positives of the \emph{Predicted aspect not in
gold set} sub-type (e.g.~extracing non-noun phrases from review texts).

For the present paper, we limited our analysis to a single dataset and
to the joint task. While this allowed us to be more focused in the
optimization efforts, it also means that the generalizability of the
findings to other datasets as well as to real-world use-cases remains a
topic for further investigation. The annotation rules of SemEval 2014
specify that only noun phrases can be aspects. However, in real-world
applications, it may be desirable to extract such aspects as well. For
example, in market research, it may be of interest to extract aspects
such as ``speed'' and ``performance'' from the sentence ``It is
extremely fast and never lags''. This would require a different
annotation scheme, and possibly a different task formulation.

Another interesting further avenue for follow-up work would be to test
the performance of a fine-tuned GPT-4 (unavailable at the time of
writing) as well as compare the performance of GPT models to that of
open source LLMs such as Llama 2 (Touvron et al. 2023b). Although the
fine-tuning appeared to significantly decrease the importance of
prompt-engineering in general, it might still be of interest to test for
example the effects of chain of thought prompting and self-correction
for performance.

In conclusion, our research demonstrates the great potential of
fine-tuned LLMs for ABSA. We found fine-tuning GPT-3.5 to the task
particularly effective, offering state-of-the-art performance at a price
point between InstructABSA and GPT-4. The performance and model size
tradeoff reflects the trend in research on transformer models: increase
in model size brings improved performance, but also increased
computational and operational costs. While our study focused on a single
benchmark dataset, it lays the foundation for broader exploration and
implementation of LLMs in ABSA across diverse datasets and use cases.

\newpage{}

\hypertarget{references}{%
\section*{References}\label{references}}
\addcontentsline{toc}{section}{References}

\hypertarget{refs}{}
\begin{CSLReferences}{1}{0}
\leavevmode\vadjust pre{\hypertarget{ref-bengio_curriculum_2009}{}}%
Bengio, Yoshua, Jérôme Louradour, Ronan Collobert, and Jason Weston.
2009. {``Curriculum Learning.''} In \emph{Proceedings of the 26th
{Annual} {International} {Conference} on {Machine} {Learning}}, 41--48.
Montreal Quebec Canada: ACM.
\url{https://doi.org/10.1145/1553374.1553380}.

\leavevmode\vadjust pre{\hypertarget{ref-DBLP:journalsux2fcorrux2fabs-2005-14165}{}}%
Brown, Tom B., Benjamin Mann, Nick Ryder, Melanie Subbiah, Jared Kaplan,
Prafulla Dhariwal, Arvind Neelakantan, et al. 2020. {``Language Models
Are Few-Shot Learners.''} \emph{CoRR} abs/2005.14165.
\url{https://arxiv.org/abs/2005.14165}.

\leavevmode\vadjust pre{\hypertarget{ref-devlin_bert_2019}{}}%
Devlin, Jacob, Ming-Wei Chang, Kenton Lee, and Kristina Toutanova. 2019.
{``{BERT}: {Pre}-Training of {Deep} {Bidirectional} {Transformers} for
{Language} {Understanding}.''} arXiv.
\url{https://doi.org/10.48550/arXiv.1810.04805}.

\leavevmode\vadjust pre{\hypertarget{ref-liu2021makes}{}}%
Liu, Jiachang, Dinghan Shen, Yizhe Zhang, Bill Dolan, Lawrence Carin,
and Weizhu Chen. 2021. {``What Makes Good in-Context Examples for
GPT-\(3\)?''} \url{https://arxiv.org/abs/2101.06804}.

\leavevmode\vadjust pre{\hypertarget{ref-mao2023gpteval}{}}%
Mao, Rui, Guanyi Chen, Xulang Zhang, Frank Guerin, and Erik Cambria.
2023. {``GPTEval: A Survey on Assessments of ChatGPT and GPT-4.''}
\url{https://arxiv.org/abs/2308.12488}.

\leavevmode\vadjust pre{\hypertarget{ref-min_rethinking_2022}{}}%
Min, Sewon, Xinxi Lyu, Ari Holtzman, Mikel Artetxe, Mike Lewis, Hannaneh
Hajishirzi, and Luke Zettlemoyer. 2022. {``Rethinking the {Role} of
{Demonstrations}: {What} {Makes} {In}-{Context} {Learning} {Work}?''}
arXiv. \url{http://arxiv.org/abs/2202.12837}.

\leavevmode\vadjust pre{\hypertarget{ref-openai2023gpt4}{}}%
OpenAI. 2023. {``GPT-4 Technical Report.''}
\url{https://arxiv.org/abs/2303.08774}.

\leavevmode\vadjust pre{\hypertarget{ref-ouyang2022training}{}}%
Ouyang, Long, Jeff Wu, Xu Jiang, Diogo Almeida, Carroll L. Wainwright,
Pamela Mishkin, Chong Zhang, et al. 2022. {``Training Language Models to
Follow Instructions with Human Feedback.''}
\url{https://arxiv.org/abs/2203.02155}.

\leavevmode\vadjust pre{\hypertarget{ref-pang-etal-2002-thumbs}{}}%
Pang, Bo, Lillian Lee, and Shivakumar Vaithyanathan. 2002. {``Thumbs up?
Sentiment Classification Using Machine Learning Techniques.''} In
\emph{Proceedings of the 2002 Conference on Empirical Methods in Natural
Language Processing ({EMNLP} 2002)}, 79--86. Association for
Computational Linguistics.
\url{https://doi.org/10.3115/1118693.1118704}.

\leavevmode\vadjust pre{\hypertarget{ref-pontiki_semeval-2014_2014}{}}%
Pontiki, Maria, Dimitris Galanis, John Pavlopoulos, Harris Papageorgiou,
Ion Androutsopoulos, and Suresh Manandhar. 2014. {``{SemEval}-2014
{Task} 4: {Aspect} {Based} {Sentiment} {Analysis}.''} In
\emph{Proceedings of the 8th {International} {Workshop} on {Semantic}
{Evaluation} ({SemEval} 2014)}, 27--35. Dublin, Ireland: Association for
Computational Linguistics. \url{https://doi.org/10.3115/v1/S14-2004}.

\leavevmode\vadjust pre{\hypertarget{ref-scaria_instructabsa_2023}{}}%
Scaria, Kevin, Himanshu Gupta, Siddharth Goyal, Saurabh Arjun Sawant,
Swaroop Mishra, and Chitta Baral. 2023. {``{InstructABSA}: {Instruction}
{Learning} for {Aspect} {Based} {Sentiment} {Analysis}.''} arXiv.
\url{https://doi.org/10.48550/arXiv.2302.08624}.

\leavevmode\vadjust pre{\hypertarget{ref-touvron2023llama}{}}%
Touvron, Hugo, Louis Martin, Kevin Stone, Peter Albert, Amjad Almahairi,
Yasmine Babaei, Nikolay Bashlykov, et al. 2023a. {``Llama 2: Open
Foundation and Fine-Tuned Chat Models.''}
\url{https://arxiv.org/abs/2307.09288}.

\leavevmode\vadjust pre{\hypertarget{ref-touvron_llama_2023}{}}%
---------, et al. 2023b. {``Llama 2: {Open} {Foundation} and
{Fine}-{Tuned} {Chat} {Models}.''} arXiv.
\url{https://doi.org/10.48550/arXiv.2307.09288}.

\leavevmode\vadjust pre{\hypertarget{ref-turney2002}{}}%
Turney, Peter D. 2002. {``Thumbs up or Thumbs down? Semantic Orientation
Applied to Unsupervised Classification of Reviews.''} In
\emph{Proceedings of the 40th Annual Meeting on Association for
Computational Linguistics}, 417--24. ACL '02. USA: Association for
Computational Linguistics.
\url{https://doi.org/10.3115/1073083.1073153}.

\leavevmode\vadjust pre{\hypertarget{ref-wang_super-naturalinstructions_2022}{}}%
Wang, Yizhong, Swaroop Mishra, Pegah Alipoormolabashi, Yeganeh Kordi,
Amirreza Mirzaei, Atharva Naik, Arjun Ashok, et al. 2022.
{``Super-{NaturalInstructions}: {Generalization} via {Declarative}
{Instructions} on 1600+ {NLP} {Tasks}.''} In \emph{Proceedings of the
2022 {Conference} on {Empirical} {Methods} in {Natural} {Language}
{Processing}}, 5085--5109. Abu Dhabi, United Arab Emirates: Association
for Computational Linguistics.
\url{https://aclanthology.org/2022.emnlp-main.340}.

\leavevmode\vadjust pre{\hypertarget{ref-yang_pyabsa_2023}{}}%
Yang, Heng, and Ke Li. 2023. {``{PyABSA}: {A} {Modularized} {Framework}
for {Reproducible} {Aspect}-Based {Sentiment} {Analysis}.''} arXiv.
\url{https://doi.org/10.48550/arXiv.2208.01368}.

\leavevmode\vadjust pre{\hypertarget{ref-zhang_sentiment_2023}{}}%
Zhang, Wenxuan, Yue Deng, Bing Liu, Sinno Jialin Pan, and Lidong Bing.
2023. {``Sentiment {Analysis} in the {Era} of {Large} {Language}
{Models}: {A} {Reality} {Check}.''} arXiv.
\url{http://arxiv.org/abs/2305.15005}.

\leavevmode\vadjust pre{\hypertarget{ref-zhang_survey_2022}{}}%
Zhang, Wenxuan, Xin Li, Yang Deng, Lidong Bing, and Wai Lam. 2022. {``A
{Survey} on {Aspect}-{Based} {Sentiment} {Analysis}: {Tasks}, {Methods},
and {Challenges}.''} \emph{IEEE Transactions on Knowledge and Data
Engineering}, 1--20. \url{https://doi.org/10.1109/TKDE.2022.3230975}.

\end{CSLReferences}

\newpage{}

\hypertarget{appendix}{%
\section{Appendix}\label{appendix}}

\hypertarget{prompts-2}{%
\subsection{Prompts}\label{prompts-2}}

\textbf{Prompt: Annotation guidelines}

Available at:
\url{https://alt.qcri.org/semeval2014/task4/data/uploads/semeval14_absa_annotationguidelines.pdf}

\textbf{Prompt: Guidelines summary}

These guidelines detail Aspect Based Sentiment Analysis Annotation for
restaurant and laptop customer reviews. The aim is to determine aspect
terms and their sentiment polarities within sentences. Aspect terms are
words or phrases describing the specific attributes of the target
entity. Sentiment polarity can be positive, negative, or neutral. For
aspect terms, annotators should mark nominal phrases explicitly
mentioning aspects and verbs or verbal names of aspects. Subjectivity
indicators, references to the target entity as a whole, and the name,
type, or model of the laptop or restaurant should not be considered as
aspect terms. Also, pronouns and implicit aspect terms should not be
annotated. For sentiment polarity, an aspect term should be classified
as positive or negative if the sentence expresses a positive or negative
opinion. Neutral polarity should be assigned when a neutral sentiment or
factual information is given about an aspect, or when the sentiment is
inferred but not explicit.

\textbf{Prompt: InstructABSA}

The output will be the aspects (both implicit and explicit), and the
aspects sentiment polarity.

\textbf{Prompt: InstructABSA with examples}

The output will be the aspects (both implicit and explicit), and the
aspects sentiment polarity.

Positive Example Example Input 1: With the great variety on the menu, I
eat here often and never get bored. Example Output 1: menu:positive

Example Input 2: Great food, good size menu, great service, and an
unpretentious setting. Example Output 2: food:positive

Negative Example Example Input 1: They did not have mayonnaise, forgot
our toast, left out ingredients (i.e., cheese in an omelet), below hot
temperatures, and the bacon was so overcooked it crumbled on the plate
when you touched it. Example Output 1: toast:negative

Example Input 2: The seats are uncomfortable if you are sitting against
the wall on wooden benches. Example Output 2: seats:negative

Neutral Example Example Input 1: I asked for a seltzer with lime, no
ice. Example Output 1: seltzer with lime:neutral

Example Input 2: They wouldn't even let me finish my glass of wine
before offering another. Example Output 2: glass of wine:neutral

Now complete the following example-input:

\textbf{Prompt: Minimal}

You extract aspects (noun phrases) and polarities (positive, neutral,
negative) from reviews

\textbf{Prompt: Reference}

SemEval 2014 Task 4: Aspect Based Sentiment Analysis

Extract aspects from customer reviews and classify their polarity.
Follow the labeling rules of the SemEval 2014 Annotation Guidelines.

\textbf{Prompt: Roleplay}

You are a perfectly accurate machine learning model for aspect based
sentiment analysis (ABSA) on reviews. Polarity labels: `positive',
`negative', `neutral'. Aspects are noun phrases that refer to specific
features of the product or service. Don't include implicit aspects.
Opinions about the product or service as a whole are not aspects. Copy
aspect terms verbatim from the reviews, don't generalize them.

\textbf{Prompt: Separate tasks}

Aspect-Based Sentiment Analysis on Customer Reviews

Your task is to perform aspect-based sentiment analysis (ABSA) on
customer reviews.

Follow the steps below to extract aspects and classify their polarity.
1. Aspect Term Extraction (ATE): - Extract all aspects mentioned in the
reviews. - Copy aspect terms verbatim, without generalizing or modifying
them. - Aspects are noun phrases that refer to specific features of the
product or service. - Don't include implicit aspects. - Opinions about
the product or service as a whole are not aspects. 2. Aspect Term
Polarity (ATP): - Classify the polarity of each aspect as `positive',
`negative', or `neutral'.

\hypertarget{in-context-examples-1}{%
\subsection{In-context examples}\label{in-context-examples-1}}

Example 1:

User: Drivers updated ok but the BIOS update froze the system up and the
computer shut down.

Assistant: {[}\{``term'': ``Drivers'', ``polarity'': ``positive''\},
\{``term'': ``BIOS update'', ``polarity'': ``negative''\}, \{``term'':
``system'', ``polarity'': ``negative''\}{]}

Example 2:

User: The food was absolutely amazing!!

Assistant: {[}\{``term'': ``food'', ``polarity'': ``positive''\}{]}

Example 3:

User: I have had to send in my laptop three times to get it fixed.

Assistant: {[}{]}

Example 4:

User: So, the menu is written in chalk above your head and it all sounds
delicious.

Assistant: {[}\{``term'': ``menu'', ``polarity'': ``neutral''\}{]}

Example 5:

User: EITHER THE COMPUTER IS TOO SLOW TO DETECT THE KEYS TYPED (THIS IS
UNLIKELY AS I AM A SLOW TYPIST) OR THE KEYBOARD SIMPLY DOES NOT DETECT
THE KEYS BEING TYPED.

Assistant: {[}\{``term'': ``KEYS'', ``polarity'': ``negative''\},
\{``term'': ``KEYBOARD'', ``polarity'': ``negative''\}, \{``term'':
``KEYS'', ``polarity'': ``negative''\}{]}

Example 6:

User: The service was bad, the food took to forever to come, we sat on
the upper level.

Assistant: {[}\{``term'': ``service'', ``polarity'': ``negative''\},
\{``term'': ``food'', ``polarity'': ``negative''\}{]}

Example 7:

User: -4 RAM slots, 2 HDD Bays*, 16GB RAM support -No Wireless Issues,
at least for me.

Assistant: {[}\{``term'': ``RAM slots'', ``polarity'': ``neutral''\},
\{``term'': ``HDD Bays'', ``polarity'': ``neutral''\}, \{``term'':
``16GB RAM support'', ``polarity'': ``neutral''\}, \{``term'':
``Wireless'', ``polarity'': ``positive''\}{]}

Example 8:

User: Definitely worth the trip to Battery Park City!

Assistant: {[}{]}

Example 9:

User: Mine came at \$1,700 w/o a DVD burner (!).

Assistant: {[}\{``term'': ``DVD burner'', ``polarity'':
``negative''\}{]}

Example 10:

User: I liked the food at this quasi-thai restaurant.

Assistant: {[}\{``term'': ``food'', ``polarity'': ``positive''\},
\{``term'': ``quasi-thai'', ``polarity'': ``positive''\}{]}

\hypertarget{json-schema}{%
\subsection{JSON schema}\label{json-schema}}

\{`name': `extract\_aspects\_and\_polarities', `description': `Extract
sentiment aspects and polarities from a text', `parameters': \{`type':
`object', `properties': \{`aspects': \{`type': `array', `description':
`An array of aspects and their polarities. If no aspects are mentioned
in the text, use an empty array.', `minItems': 0, `items': \{`type':
`object', `properties': \{`term': \{`type': `string', `description': `An
aspect term, which is a verbatim text snippet. Single or multiword terms
naming particular aspects of the reviewed product or service.'\},
`polarity': \{`type': `string', `enum': {[}`positive', `neutral',
`negative'{]}, `description': ``The polarity expressed towards the
aspect term. Valid polarities are `positive', `neutral',
`negative'.''\}\}, `required': {[}`term', `polarity'{]}\}\}\},
`required': {[}`aspects'{]}\}\}

\end{document}